\documentclass[11pt]{article}
\usepackage{authblk}
\usepackage{eacl2017}
\usepackage{times}
\usepackage{latexsym}
\usepackage{amsmath}
\usepackage{graphicx}
\graphicspath{ {images/} }
\usepackage{multirow}
\usepackage{multicol}
\usepackage{hhline}
\usepackage{url}
\usepackage{color}
\usepackage[font=footnotesize,labelfont=bf]{caption}
\usepackage{tikz}
\usepackage{tabularx}
\usepackage{tikz-dependency}
\usepackage{footnote}
\usepackage{enumitem}

\makesavenoteenv{tabular}

\newcommand{\specialcell}[2][c]{%
  \begin{tabular}[#1]{@{}c@{}}#2\end{tabular}}

\newcommand\smalltt[1]{\texttt{\small #1}}

\newcommand{\mediumfont}{\fontsize{10.5pt}{11.5pt}\selectfont}

\eaclfinalcopy 
\def\eaclpaperid{41} 

\title{Hypernyms under Siege:\\Linguistically-motivated Artillery for Hypernymy Detection}

\author[1]{\bf Vered Shwartz}
\author[2,3]{\bf Enrico Santus}
\author[4]{\bf Dominik Schlechtweg}
\affil[1]{Bar-Ilan University\\Ramat-Gan, Israel}
\affil[2]{Singapore University of Technology and Design\\Singapore}
\affil[3]{The Hong Kong Polytechnic University\\Hong Kong}
\affil[4]{University of Stuttgart\\Stuttgart, Germany}
\affil[  ]{\mediumfont\tt  \{vered1986,esantus\}@gmail.com,~~dominik.schlechtweg@gmx.de}

\begin{document}

\maketitle

\begin{abstract}
The fundamental role of hypernymy in NLP has motivated the development of many methods for the automatic identification of this relation, most of which rely on word distribution. 
We investigate an extensive number of such unsupervised measures, using several distributional semantic models that differ by context type and feature weighting. We analyze the performance of the different methods based on their linguistic motivation.
Comparison to the state-of-the-art supervised methods shows that while supervised methods generally outperform the unsupervised ones, the former are sensitive to the distribution of training instances, hurting their reliability. Being based on general linguistic hypotheses and independent from training data, unsupervised measures are more robust, and therefore are still useful artillery for hypernymy detection.
\end{abstract}

\section{Introduction} 

In the last two decades, the NLP community has invested a consistent effort in developing automated methods to recognize hypernymy. Such effort is motivated by the role this semantic relation plays in a large number of tasks, such as taxonomy creation \cite{snow2006semantic,navigli2011graph} and recognizing textual entailment \cite{dagan2013recognizing}. The task has appeared to be, however, a challenging one, and the numerous approaches proposed to tackle it have often shown limitations.

Early corpus-based methods have exploited patterns that may indicate hypernymy (e.g. ``\emph{animals} such as \emph{dogs}'') \cite{hearst1992automatic,snow2004learning}, but the recall limitation of this approach, requiring both words to co-occur in a sentence, motivated the development of methods that rely on adaptations of the \emph{distributional hypothesis} \cite{harris1954distributional}. 

The first distributional approaches were unsupervised, assigning a score for each $(x,y)$ word-pair, which is expected to be higher for hypernym pairs than for negative instances. Evaluation is performed using ranking metrics inherited from information retrieval, such as Average Precision (AP) and Mean Average Precision (MAP). Each measure exploits a certain linguistic hypothesis such as the \emph{distributional inclusion hypothesis} \cite{weeds2003general,kotlerman2010directional} and the \emph{distributional informativeness hypothesis} \cite{santus2014chasing,rimell2014distributional}. 

In the last couple of years, the focus of the research community shifted to supervised distributional methods, in which each $(x,y)$ word-pair is represented by a combination of $x$ and $y$'s word vectors (e.g. concatenation or difference), and a classifier is trained on these resulting vectors to predict hypernymy \cite{baroni2012entailment,roller2014inclusive,weeds2014learning}. While the original methods were based on count-based vectors, in recent years they have been used with word embeddings \cite{mikolov2013distributed,pennington2014glove}, and have gained popularity thanks to their ease of use and their high performance on several common datasets. However, there have been doubts on whether they can actually learn to recognize hypernymy \cite{levy2015supervised}.

Additional recent hypernymy detection methods include a multimodal perspective \cite{kiela2015exploiting}, a supervised method using unsupervised measure scores as features \cite{santus2016nine}, and a neural method integrating path-based and distributional information \cite{shwartz2016improving}. 

In this paper we perform an extensive evaluation of various unsupervised distributional measures for hypernymy detection, using several distributional semantic models that differ by context type and feature weighting. Some measure variants and context-types are tested for the first time.\footnote{Our code and data are available at:\\\scriptsize{\url{https://github.com/vered1986/UnsupervisedHypernymy}}}

We demonstrate that since each of these measures captures a different aspect of the hypernymy relation, there is no single measure that consistently performs well in discriminating hypernymy from different semantic relations. We analyze the performance of the measures in different settings and suggest a principled way to select the suitable measure, context type and feature weighting according to the task setting, yielding consistent performance across datasets. 

We also compare the unsupervised measures to the state-of-the-art supervised methods. We show that supervised methods outperform the unsupervised ones, while also being more efficient, computed on top of low-dimensional vectors. At the same time, however, our analysis reassesses previous findings suggesting that supervised methods do not actually learn the relation between the words, but only characteristics of a single word in the pair \cite{levy2015supervised}. Moreover, since the features in embedding-based classifiers are latent, it is difficult to tell what the classifier has learned. We demonstrate that unsupervised methods, on the other hand, do account for the relation between words in a pair, and are easily interpretable, being based on general linguistic hypotheses.

\section{Distributional Semantic Spaces}
\label{sec:dsm}

We created multiple distributional semantic spaces that differ in their context type and feature weighting. 
As an underlying corpus we used a concatenation of the following two corpora: \smalltt{ukWaC} \cite{ferraresi2007building}, a 2-billion word corpus constructed by crawling the .uk domain, and \smalltt{WaCkypedia\_EN} \cite{baroni2009wacky}, a 2009 dump of the English Wikipedia. Both corpora include POS, lemma and dependency parse annotations. Our vocabulary (of target and context words) includes only nouns, verbs and adjectives that occurred at least 100 times in the corpus.

\begin{figure}[t]
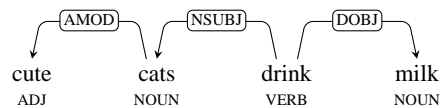
 
\centering
\small
\begin{dependency}
\begin{deptext}[column sep=3em]
		cute \& cats \& drink \& milk\\
		\tiny ADJ \& \tiny NOUN \& \tiny VERB \& \tiny NOUN \\
  \end{deptext}
	\depedge{2}{1}{AMOD}
  \depedge{3}{2}{NSUBJ}
  \depedge{3}{4}{DOBJ}
\end{dependency}
\vspace*{-10pt}
\caption{An example dependency tree of the sentence \emph{cute cats drink milk}, with the target word \emph{cats}. 
The dependency-based contexts are \emph{drink-v:nsubj} and \emph{cute-a:amod$^{-1}$}. 
The joint-dependency context is \emph{drink-v\#milk-n}. Differently from \newcite{chersoni2016joint}, we exclude the dependency tags to mitigate the sparsity of contexts.}
\label{fig:dep}
\vspace*{-10pt}
\end{figure}

\paragraph{Context Type} We use several context types: 

\begin{itemize}[leftmargin=*]

\vspace*{-2pt}
\item \textbf{Window-based contexts:} the contexts of a target word $w_i$ are the words surrounding it in a $k$-sized window: $w_{i-k}, ..., w_{i-1}, w_{i+1}, ..., w_{i+k}.$\\If the context-type is directional, words occurring before and after $w_i$ are marked differently, i.e.: $w_{i-k}/l, ..., w_{i-1}/l, w_{i+1}/r, ..., w_{i+k}/r$.\\Out-of-vocabulary words are filtered out before applying the window. We experimented with window sizes 2 and 5, directional and indirectional (\smalltt{win2, win2d, win5, win5d}). 

\vspace*{-2pt}
\item \textbf{Dependency-based contexts:} rather than adjacent words in a window, we consider neighbors in a dependency parse tree \cite{pado2007dependency,baroni2010distributional}. The contexts of a target word $w_i$ are its parent and daughter nodes in the dependency tree (\smalltt{dep}). We also experimented with a joint dependency context inspired by \newcite{chersoni2016joint}, in which the contexts of a target word are the parent-sister pairs in the dependency tree (\smalltt{joint}). See Figure~\ref{fig:dep} for an illustration. 
\end{itemize}
\vspace*{-2pt}

\paragraph{Feature Weighting} Each distributional semantic space is spanned by a matrix $M$ in which each row corresponds to a target word while each column corresponds to a context. The value of each cell $M_{i,j}$ represents the association between the target word $w_i$ and the context $c_j$. We experimented with two feature weightings:

\begin{itemize}[leftmargin=*]
\vspace*{-2pt}
\item \textbf{Frequency} - raw frequency (no weighting): $M_{i,j}$ is the number of co-occurrences of $w_i$ and $c_j$ in the corpus.

\vspace*{-2pt}
\item \textbf{Positive PMI (PPMI)} - pointwise mutual information (PMI) \cite{church1990word} is defined as the log ratio between the joint probability of $w$ and $c$ and the product of their marginal probabilities: $PMI(w, c) = log\frac{\hat{P}(w,c)}{\hat{P}(w)\hat{P}(c)}$, where $\hat{P}(w), \hat{P}(c)$, and $\hat{P}(w,c)$ are estimated by the relative frequencies of a word $w$, a context $c$ and a word-context pair $(w,c)$, respectively. To handle unseen pairs $(w,c)$, yielding $PMI(w,c) = log(0) = -\infty$, PPMI \cite{bullinaria2007extracting} assigns zero to negative PMI scores: $PPMI(w,c) = max(PMI(w,c), 0)$.

\end{itemize}

\vspace*{-2pt}
In addition, one of the measures we used \cite{santus2014chasing} required a third feature weighting:

\begin{itemize}
\vspace*{-2pt}
\item \textbf{Positive LMI (PLMI)} - positive local mutual information (PLMI) \cite{evert2005statistics,evert2008corpora}. PPMI was found to have a bias towards rare events. PLMI simply balances PPMI by multiplying it by the co-occurrence frequency of $w$ and $c$: $PLMI(w,c) = freq(w,c) \cdot PPMI(w,c)$.

\end{itemize}
\vspace*{-2pt}

\setlength{\abovedisplayskip}{0.01cm}
\setlength{\belowdisplayskip}{0.01cm}

\section{Unsupervised Hypernymy Detection Measures}
\label{sec:measures}

We experiment with a large number of unsupervised measures proposed in the literature for distributional hypernymy detection, with some new variants. In the following section, $\vec{v}_x$ and $\vec{v}_y$ denote $x$ and $y$'s word vectors (rows in the matrix $M$). We consider the scores as measuring to what extent $y$ is a hypernym of $x$ ($x \rightarrow y$). 

\newcommand{\norm}[1]{\left\lVert #1 \right\rVert}

\subsection{Similarity Measures}
\label{sec:similarity_measures}

Following the \emph{distributional hypothesis} \cite{harris1954distributional}, similar words share many contexts, thus have a high similarity score. Although the hypernymy relation is asymmetric, similarity is one of its properties \cite{santus2014chasing}.

\begin{itemize}[leftmargin=*]

	\item \textbf{Cosine Similarity} \cite{salton1986introduction} A symmetric similarity measure:
	
	\begin{equation*}
	cos(x,y) = \frac{\vec{v}_x \cdot \vec{v}_y}{\norm{\vec{v}_x} \cdot \norm{\vec{v}_y}}
	\end{equation*}
	
	\item \textbf{Lin Similarity} \cite{lin1998information} A symmetric similarity measure that quantifies the ratio of shared contexts to the contexts of each word:
	
	\begin{equation*}
	Lin(x,y)= \frac{\Sigma_{c \in \vec{v}_x \cap \vec{v}_y}{[\vec{v}_x[c] + \vec{v}_y[c]]}}{\Sigma_{c \in \vec{v}_x }{\vec{v}_x[c]} + \Sigma_{c \in \vec{v}_y}{\vec{v}_y[c]}}
	\end{equation*}
	
	\item \textbf{APSyn} \cite{santus2016unsupervised} A symmetric measure that computes the extent of intersection among the $N$ most related contexts of two words, weighted according to the rank of the shared contexts (with $N$ as a hyper-parameter):

	\begin{equation*}
	\hspace*{-7pt}
	APSyn(x, y) = \Sigma_{c \in N(\vec{v}_x) \cap N(\vec{v}_y)}{\frac{1}{\frac{rank_x(c) + rank_y(c)}{2}}}
	\end{equation*}
	
\end{itemize}

\subsection{Inclusion Measures}
\label{sec:inclusion_measures}

According to the \emph{distributional inclusion hypothesis}, the prominent contexts of a hyponym ($x$) are expected to be included in those of its hypernym ($y$). 

\begin{itemize}[leftmargin=*]

	\item \textbf{Weeds Precision} \cite{weeds2003general} A directional precision-based similarity measure. This measure quantifies the weighted inclusion of $x$'s contexts by $y$'s contexts: 
	\begin{equation*}
	WeedsPrec(x \rightarrow y)= \frac{\Sigma_{c \in \vec{v}_x \cap \vec{v}_y}{\vec{v}_x[c]}}{\Sigma_{c \in \vec{v}_x}{\vec{v}_x[c]}}
	\end{equation*}
	
	\item \textbf{cosWeeds} \cite{lenci2012identifying} Geometric mean of cosine similarity and Weeds precision:
	\small
	\begin{equation*}
	cosWeeds(x \rightarrow y) = \sqrt{cos(x,y) \cdot WeedsPrec(x \rightarrow y)}
	\end{equation*}
	\normalsize
	
	\item \textbf{ClarkeDE} \cite{clarke2009context} Computes degree of inclusion, by quantifying weighted coverage of the hyponym's contexts by those of the hypernym:
	\small
	\begin{equation*}
	CDE(x \rightarrow y) = \frac{\Sigma_{c \in \vec{v}_x \cap \vec{v}_y}{min(\vec{v}_x[c], \vec{v}_y[c])}}{\Sigma_{c \in \vec{v}_x}{\vec{v}_x[c]}}
	\end{equation*}
	\normalsize

	\item \textbf{balAPinc} \cite{kotlerman2010directional} Balanced average precision inclusion.
	\small
	\begin{equation*}
	APinc(x \rightarrow y) = \frac{\sum_{r=1}^{N_y}{[P(r) \cdot rel(c_r)]}}{N_y}
	\end{equation*}
	\normalsize
	\noindent is an adaptation of the average precision measure from information retrieval for the inclusion hypothesis. $N_y$ is the number of non-zero contexts of $y$ and $P(r)$ is the precision at rank $r$, defined as the ratio of shared contexts with $y$ among the top $r$ contexts of $x$. $rel(c)$ is the relevance of a context $c$, set to 0 if $c$ is not a context of $y$, and to $1-\frac{rank_y(c)}{N_y+1}$ otherwise, where $rank_y(c)$ is the rank of the context $c$ in $y$'s sorted vector. Finally, 
	\small
	\begin{equation*}
	balAPinc(x \rightarrow y) = \sqrt{Lin(x,y) \cdot APinc(x \rightarrow y)}
	\end{equation*}
	\normalsize
	\noindent is the geometric mean of APinc and Lin similarity.

	\item \textbf{invCL} \cite{lenci2012identifying} Measures both distributional inclusion of $x$ in $y$ and distributional
non-inclusion of $y$ in $x$: 

	\small
	\begin{equation*}
	invCL(x \rightarrow y) = \sqrt{CDE(x \rightarrow y) \cdot (1-CDE(y \rightarrow x))}
	\end{equation*}
	\normalsize
	
	\end{itemize}
	
\subsection{Informativeness Measures}
\label{sec:informativeness_measures}

According to the \emph{distributional informativeness hypothesis}, hypernyms tend to be less informative than hyponyms, as they are likely to occur in more general contexts than their hyponyms.

\begin{itemize}[leftmargin=*]

	\item \textbf{SLQS} \cite{santus2014chasing} 
	\begin{equation*}
	SLQS(x \rightarrow y) = 1 - \frac{E_x}{E_y}
	\end{equation*}
	The informativeness of a word $x$ is evaluated as the median entropy of its top $N$ contexts: $E_x = median_{i=1}^N(H(c_i))$, where $H(c)$ is the entropy of context c. 
	
	\item \textbf{SLQS Sub} A new variant of SLQS based on the assumption that if $y$ is judged to be a hypernym of $x$ to a certain extent, then $x$ should be judged to be a hyponym of $y$ to the same extent (which is not the case for regular SLQS). This is achieved by subtraction: 
	\begin{equation*}
	SLQS_{sub}(x \rightarrow y) = E_y - E_x
	\end{equation*}
	
	\noindent It is weakly symmetric in the sense that $SLQS_{sub}(x \rightarrow y) = - SLQS_{sub}(y \rightarrow x)$.
	
	\item[] SLQS and SLQS Sub have 3 hyper-parameters: i) the number of contexts $N$; ii) whether to use median or average entropy among the top $N$ contexts;
	and iii) the feature weighting used to sort the contexts by relevance (i.e., PPMI or PLMI).
	
	\item \textbf{SLQS Row} Differently from SLQS, SLQS Row computes the entropy of the target rather than the average/median entropy of the contexts, as an alternative way to compute the generality of a word.\footnote{In our preliminary experiments, we noticed that the entropies of the targets and those of the contexts are not highly correlated, yielding a Spearman's correlation of up to 0.448 for window based spaces, and up to 0.097 for the dependency-based ones ($p<0.01$).} In addition, parallel to SLQS we tested SLQS Row with subtraction, \textbf{SLQS Row Sub}.
	
	\item \textbf{RCTC} \cite{rimell2014distributional} Ratio of change in topic coherence: 
	\small
	\begin{equation*}
	RCTC(x \rightarrow y)=\frac{TC(\text{t}_x)/TC(\text{t}_{x \backslash y})}{TC(\text{t}_y)/TC(\text{t}_{y \backslash x})}
	\end{equation*}
	\normalsize
	\noindent where t$_x$ are the top $N$ contexts of $x$, considered as $x$'s \emph{topic}, and t$_{x \backslash y}$ are the top $N$ contexts of $x$ which are not contexts of $y$. $TC(A)$ is the topic coherence of a set of words $A$, defined as the median pairwise PMI scores between words in $A$. $N$ is a hyper-parameter. The measure is based on the assumptions that excluding $y$'s contexts from $x$'s increases the coherence of the topic, while excluding $x$'s contexts from $y$'s decreases the coherence of the topic. We include this measure under the informativeness inclusion, as it is based on a similar hypothesis.
	
\end{itemize}

\subsection{Reversed Inclusion Measures}
\label{sec:rev_inclusion_measures}

These measures are motivated by the fact that, even though---being more general---hypernyms are expected to occur in a larger set of contexts, sentences like ``the \emph{vertebrate} barks'' or ``the \emph{mammal} arrested the thieves'' are not common, since hyponyms are more specialized and are hence more appropriate in such contexts. On the other hand, hyponyms are likely to occur in broad contexts (e.g. \emph{eat}, \emph{live}), where hypernyms are also appropriate. In this sense, we can define the \emph{reversed inclusion hypothesis}: ``hypernym's contexts are likely to be included in the hyponym's contexts''. The following variants are tested for the first time.

\begin{itemize}[leftmargin=*]

\item \textbf{Reversed Weeds} 
\begin{equation*}
RevWeeds(x \rightarrow y)= Weeds(y \rightarrow x)
\end{equation*}

\item \textbf{Reversed ClarkeDE} 
\begin{equation*}
RevCDE(x \rightarrow y) = CDE(y \rightarrow x)
\end{equation*}
	
\end{itemize}

\section{Datasets}
\label{sec:datasets}

\begin{table}
\center
\small
\begin{tabular}{ | c | c | c | c | }
    \hline
    \textbf{dataset} & \specialcell{\textbf{relations}} & \textbf{\#instances} & \textbf{size} \\ \hline
		\multirow{8}{*}{BLESS} 	& hypernym & 1,337 & \multirow{8}{*}{26,554} \\ \hhline{~--~}
														& meronym & 2,943 & \\ \hhline{~--~}
														& coordination & 3,565 & \\ \hhline{~--~}
														& event & 3,824 & \\ \hhline{~--~}
														& attribute & 2,731 & \\ \hhline{~--~} 
														& random-n & 6,702 & \\ \hhline{~--~}
														& random-v & 3,265 & \\ \hhline{~--~}
														& random-j & 2,187 & \\ \hline 
		\multirow{5}{*}{EVALution} 	& hypernym & 3,637 & \multirow{5}{*}{13,465\footnotemark} \\ \hhline{~--~}
																& meronym & 1,819 & \\ \hhline{~--~}
																& attribute & 2,965 & \\ \hhline{~--~} 
																& synonym & 1,888 & \\ \hhline{~--~}
																& antonym & 3,156 & \\ \hline 
		\multirow{3}{*}{Lenci/Benotto} 	& hypernym & 1,933 & \multirow{3}{*}{5,010} \\ \hhline{~--~}
																		& synonym & 1,311 & \\ \hhline{~--~}
																		& antonym & 1,766 & \\ \hline 
		\multirow{2}{*}{Weeds} 	& hypernym & 1,469 & \multirow{2}{*}{2,928} \\ \hhline{~--~}
														& coordination & 1,459 & \\ \hline 
\end{tabular}
\vspace{-7pt}
	\caption{The semantic relations, number of instances in each relation, and size of each dataset.}
	\label{tab:datasets}
	\vspace{-10pt}
\end{table}
\footnotetext{We removed the \emph{entailment} relation, which had too few instances, and conflated relations to coarse-grained relations (e.g. \emph{HasProperty} and \emph{HasA} into \emph{attribute}).}

\begin{table*}
	\center
	\small
	\hspace*{-10pt}
	\begin{tabular}{ | c | c | c | c | c | c | c | c |}
		\hline
		\textbf{dataset} & \textbf{hyper vs. relation} & \specialcell{\textbf{measure}} & \specialcell{\textbf{context}\\\textbf{type}}
		& \specialcell{\textbf{feature}\\\textbf{weighting}} & \specialcell{\textbf{hyper-parameters}}  & 
		\textbf{AP@100} & \textbf{AP@All} \\ \hline
		\multirow{9}{*}{\textbf{EVALution}} & \multirow{1}{*}{all other relations} & invCL & joint & freq & - & 0.661 & 0.353 \\ \hhline{~-------}
		& \multirow{1}{*}{meronym} & APSyn & joint & freq & \specialcell{\emph{N=500}} & 0.883 & 0.675
		  \\ \hhline{~-------}
		& \multirow{1}{*}{attribute} & APSyn & joint & freq & \specialcell{\emph{N=500}} & 0.88  & 0.651
		  \\ \hhline{~-------}
		& \multirow{3}{*}{antonym} & \multirow{3}{*}{SLQS\_row} & joint & freq & \multirow{3}{*}{\emph{-}}  & 0.74   & 0.54
		  \\ \hhline{~~~--~--}
		& & & joint & ppmi &  & 0.74   & 0.55
		  \\ \hhline{~~~--~--}
		& & & joint & plmi & & 0.74   & 0.537
		  \\ \hhline{~-------}
		& \multirow{3}{*}{synonym} & \multirow{3}{*}{SLQS\_row} & joint & freq & \multirow{3}{*}{\emph{-}} & 0.83   & 0.647
		  \\ \hhline{~~~--~--}
		& &  & joint & ppmi & & 0.83  & 0.657
		  \\ \hhline{~~~--~--}
		& &  & joint & plmi & & 0.83 & 0.645
		  \\ \hline \hline
		\multirow{8}{*}{\textbf{BLESS}} & \multirow{1}{*}{all other relations} & invCL & win5 & freq & - & 0.54 & 0.051 \\ \hhline{~-------}
		& \multirow{2}{*}{meronym} & SLQS$_{sub}$ & win5d & freq & \specialcell{\emph{N=100, median, plmi}} & 1.0   & 0.76
		  \\ \hhline{~~------}
		& & SLQS & win5d & freq & \specialcell{\emph{N=100, median, plmi}} & 1.0  & 0.758
		  \\ \hhline{~-------}
		& \multirow{1}{*}{coord} & SLQS$_{sub}$ & joint & freq & \specialcell{\emph{N=50, average, plmi}} & 0.995  & 0.537
		  \\ \hhline{~-------}
		& \multirow{2}{*}{attribute} & SLQS$_{sub}$ & dep & plmi & \specialcell{\emph{N=70, average, plmi}} & 1.0 & 0.74
		\\ \hhline{~~------}
		& & cosine & joint & freq & \specialcell{\emph{-}} & 1.0 & 0.622
		  \\ \hhline{~-------}
		& \multirow{1}{*}{event} & APSyn & dep & freq & \specialcell{\emph{N=1000}} & 1.0 & 0.779
		   \\ \hline
		\multirow{3}{*}{\specialcell{\textbf{Lenci/}\\\textbf{Benotto}}} & \multirow{1}{*}{all other relations} & APSyn & joint & freq & \specialcell{\emph{N=1000}} & 0.617 & 0.382 \\ \hhline{~-------}
		& \multirow{1}{*}{antonym} & APSyn & dep & freq & \specialcell{\emph{N=1000}} & 0.861  & 0.624
		 \\ \hhline{~-------}
		& \multirow{1}{*}{synonym} & SLQS\_row$_{sub}$ & joint & ppmi & \specialcell{\emph{-}} & 0.948  & 0.725
		\\ \hline \hline
		\multirow{2}{*}{\textbf{Weeds}} & \multirow{1}{*}{all other relations} & clarkeDE & win5d & freq & \specialcell{\emph{-}}  & 0.911  & 0.441 \\ \hhline{~-------}
		& \multirow{1}{*}{coord} & clarkeDE & win5d & freq & \specialcell{\emph{-}}  & 0.911  & 0.441 \\ \hline
	\end{tabular}
	\vspace*{-7pt}
	\caption{Best performing unsupervised measures on each dataset in terms of Average Precision (AP) at $k=100$, for hypernym vs. all other relations and vs. each single relation. AP for $k=all$ is also reported for completeness. We excluded the experiments of hypernym vs. random-(n, v, j) for brevity; most of the similarity and some of the inclusion measures achieve $AP@100=1.0$ in these experiments.}
	\label{tab:unsupervised}
\end{table*}

We use four common semantic relation datasets: BLESS \cite{baroni2011we}, EVALution \cite{santus2015evalution}, Lenci/Benotto \cite{benotto2015distributional}, and Weeds \cite{weeds2014learning}. The datasets were constructed either using knowledge resources (e.g. WordNet, Wikipedia), crowd-sourcing or both. The semantic relations and the size of each dataset are detailed in Table~\ref{tab:datasets}.

In our distributional semantic spaces, a target word is represented by the word and its POS tag. While BLESS and Lenci/Benotto contain this information, we needed to add POS tags to the other datasets. For each pair $(x, y)$, we considered 3 pairs $(x\text{-}p, y\text{-}p)$ for $p \in \{noun, adjective, verb\}$, and added the respective pair to the dataset only if the words were present in the corpus.\footnote{Lenci/Benotto includes pairs to which more than one relation is assigned, e.g. when $x$ or $y$ are polysemous, and related differently in each sense. We consider $y$ as a hypernym of $x$ if hypernymy holds in some of the words' senses. Therefore, when a pair is assigned both hypernymy and another relation, we only keep it as hypernymy.}

We split each dataset randomly to 90\% test and 10\% validation. The validation sets are used to tune the hyper-parameters of several measures: SLQS (Sub), APSyn and RCTC.

\section{Experiments}
\label{sec:experiments}

\subsection{Comparing Unsupervised Measures}
\label{sec:unsupervised}

\begin{table*}
\center
\small
\begin{tabular}{ | c | c | c | c | }
    \hline
		\textbf{relation} & \specialcell{\textbf{measure}} & \specialcell{\textbf{context type}} & \specialcell{\textbf{feature weighting}} \\ \hline\hline		
		\multirow{3}{*}{\textbf{meronym}} & cosWeeds & dep & ppmi \\ \hhline{~---}
																			& Weeds & dep / joint & ppmi \\ \hhline{~---}
																			& ClarkeDE & dep / joint & ppmi / freq \\ \hline
		\multirow{4}{*}{\textbf{attribute}} & APSyn & joint & freq \\ \hhline{~---}
																				& cosine & joint & freq \\ \hhline{~---} 
																				& Lin & dep & ppmi \\ \hhline{~---}
																				& cosine & dep & ppmi \\ \hline
		\textbf{antonym} & SLQS & - & - \\ \hline 
		\multirow{3}{*}{\textbf{synonym}} & SLQS\_row & joint & (freq/ppmi/plmi) \\ \hhline{~---}
																			& SLQS\_row/SLQS\_row\_sub & dep & ppmi \\ \hhline{~---} 
																			& invCL & win2/5/5d & freq \\ \hline
		\textbf{coordination} & \multicolumn{3}{|c|}{-} \\ \hline
\end{tabular}
\vspace*{-7pt}
	\caption{Intersection of datasets' top-performing measures when discriminating between hypernymy and each other relation.}
	\label{tab:best_measure_for_relation}
\end{table*}

In order to evaluate the unsupervised measures described in Section~\ref{sec:measures}, we compute the measure scores for each $(x, y)$ pair in each dataset. 
We first measure the method's ability to discriminate hypernymy from all other relations in the dataset, i.e. by considering hypernyms as positive instances, and other word pairs as negative instances. In addition, we measure the method's ability to discriminate hypernymy from every other relation in the dataset by considering \emph{one} relation at a time. For a relation $\mathcal{R}$ we consider only $(x, y)$ pairs that are annotated as either hypernyms (positive instances) or $\mathcal{R}$ (negative instances). We rank the pairs according to the measure score and compute average precision (AP) at $k=100$ and $k=all$.\footnote{We tried several cut-offs and chose the one that seemed to be more informative in distinguishing between the unsupervised measures.}

Table~\ref{tab:unsupervised} reports the best performing measure(s), with respect to $AP@100$, for each relation in each dataset. The first observation is that there is no single combination of measure, context type and feature weighting that performs best in discriminating hypernymy from all other relations. In order to better understand the results, we focus on the second type of evaluation, in which we discriminate hypernyms from each other relation. 

The results show preference to the syntactic context-types (\smalltt{dep} and \smalltt{joint}), which might be explained by the fact that these contexts are richer (as they contain both proximity and syntactic information) and therefore more discriminative. In feature weighting there is no consistency, but interestingly, raw frequency appears to be successful in hypernymy detection, contrary to previously reported results for word similarity tasks, where PPMI was shown to outperform it \cite{bullinaria2007extracting,levy2015improving}. 

The new SLQS variants are on top of the list in many settings. In particular they perform well in discriminating hypernyms from symmetric relations (antonymy, synonymy, coordination). 

The measures based on the \emph{reversed inclusion hypothesis} performed inconsistently, achieving perfect score in the discrimination of hypernyms from unrelated words, and performing well in few other cases, always in combination with syntactic contexts.

Finally, the results show that there is no single combination of measure and parameters that performs consistently well for all datasets and classification tasks. In the following section we analyze the best combination of measure, context type and feature weighting to distinguish hypernymy from any other relation.

\subsection{Best Measure Per Classification Task}
\label{sec:measure_per_relation}

We considered all relations that occurred in two datasets. For such relation, for each dataset, we ranked the measures by their AP@100 score, selecting those with score $\geq 0.8$.\footnote{We considered at least 10 measures, allowing scores slightly lower than 0.8 when others were unavailable.} Table~\ref{tab:best_measure_for_relation} displays the intersection of the datasets' best measures.

\begin{table*}
\hspace*{-10pt}
	\center
	\small
	\begin{tabular}{ | c | c || c | c | c | c || c | c | c | c | }
		\hline
		\multirow{3}{*}{\textbf{dataset}} & \multirow{3}{*}{\specialcell{\textbf{hyper vs.} \\ \textbf{relation}}} & 
		\multicolumn{4}{|c||}{\specialcell{\textbf{best supervised}}} & 
		\multicolumn{4}{|c|}{\specialcell{\textbf{best unsupervised}}} \\ \hhline{~~--------}
		& & \textbf{method} & \textbf{vectors} & \textbf{penalty} & \specialcell{\textbf{AP} \\ \textbf{@100}} & 
		\specialcell{\textbf{measure}} & \specialcell{\textbf{context}\\\textbf{type}} & \specialcell{\textbf{feature}\\\textbf{weighting}} & 
		\specialcell{\textbf{AP} \\ \textbf{@100}} \\ \hline \hline
		\multirow{4}{*}{\textbf{EVALution}} & 
		\multirow{1}{*}{meronym} & concat & dep-based & $L_2$ & 0.998 & APSyn & joint & freq	& 0.886 \\ \hhline{~---------}
		& \multirow{1}{*}{attribute} & concat & Glove-100 & $L_2$	& 1.000 &	invCL & dep & ppmi & 0.877 \\ \hhline{~---------}
		& \multirow{1}{*}{antonym} & concat & dep-based & $L_2$	& 1.000	& invCL & joint & ppmi & 0.773 \\ \hhline{~---------}
		& \multirow{1}{*}{synonym} & concat & dep-based & $L_1$	& 0.996	& SLQS$_{sub}$ & win2 & plmi & 0.813 \\ \hline \hline
		\multirow{7}{*}{\textbf{BLESS}} & 
		\multirow{1}{*}{meronym} & concat & Glove-50 & $L_1$ &	1.000 & SLQS$_{sub}$ & win5 & freq & 0.939 \\ \hhline{~---------}
		& \multirow{1}{*}{coord} & concat & Glove-300 & $L_1$	& 1.000 & SLQS\_row$_{sub}$ & joint & plmi & 0.938  \\ \hhline{~---------}
		& \multirow{1}{*}{attribute} & concat & Glove-100 & $L_1$	& 1.000	& SLQS$_{sub}$ & dep & freq	& 0.938	\\ \hhline{~---------}
		& \multirow{1}{*}{event} & concat & Glove-100 & $L_1$ & 1.000 & SLQS$_{sub}$ & dep & freq & 0.847	\\ \hhline{~---------}
		& \multirow{1}{*}{random-n} & concat & word2vec & $L_1$	& 0.995	& cosWeeds & win2d & ppmi & 0.818	\\ \hhline{~---------}
		& \multirow{1}{*}{random-j} & concat & Glove-200 & $L_1$ & 1.000	& SLQS$_{sub}$ & dep & freq	& 0.917	\\ \hhline{~---------}
		& \multirow{1}{*}{random-v} & concat & word2vec & $L_1$ & 1.000 & SLQS$_{sub}$ & dep & freq & 0.895	\\ \hline \hline
		\multirow{2}{*}{\specialcell{\textbf{Lenci/}\\\textbf{Benotto}}} & 
		\multirow{1}{*}{antonym} & concat & dep-based & $L_2$ & 0.917	& invCL & joint & ppmi & 0.807 \\ \hhline{~---------}
		& \multirow{1}{*}{synonym} & concat & Glove-300 & $L_1$ & 0.946 & invCL & win5d & freq	& 0.914	\\ \hline \hline
		\multirow{2}{*}{\textbf{Weeds}} & 
		\multirow{2}{*}{coord} & \multirow{2}{*}{concat} & \multirow{2}{*}{dep-based} & \multirow{2}{*}{$L_2$} & \multirow{2}{*}{0.873} 
		& invCL & win2d & freq & \multirow{2}{*}{0.824} \\ \hhline{~~~~~~---~}
		& & & & & & SLQS\_row$_{sub}$ & joint & ppmi & \\ \hline
	\end{tabular}
	\vspace*{-8pt}
	\caption{Best performance on the validation set (10\%) of each dataset for the supervised and unsupervised measures, in terms of Average Precision (AP) at $k=100$, for hypernym vs. each single relation.}
	\label{tab:validation_results}
	\vspace*{-13pt}
\end{table*}

\paragraph{Hypernym vs. Meronym} The inclusion hypothesis seems to be most effective in discriminating between hypernyms and meronyms under syntactic contexts.
We conjecture that the window-based contexts are less effective since they capture topical context words, that might be shared also among holonyms and their meronyms (e.g. \emph{car} will occur with many of the neighbors of \emph{wheel}). However, since meronyms and holonyms often have different functions, their functional contexts, which are expressed in the syntactic context-types, are less shared. This is where they mostly differ from hyponym-hypernym pairs, which are of the same function (e.g. \emph{cat} is a type of \emph{animal}).

Table~\ref{tab:unsupervised} shows that SLQS performs well in this task on BLESS. This is contrary to previous findings that suggested that SLQS is weak in discriminating between hypernyms and meronyms, as in many cases the holonym is more general than the meronym \cite{shwartz2016improving}.\footnote{In the hypernymy dataset of \newcite{shwartz2016improving}, nearly 50\% of the SLQS false positive pairs were meronym-holonym pairs, in many of which the holonym is more general than the meronym by definition, e.g. \emph{(mauritius, africa)}.
} The surprising result could be explained by the nature of meronymy in this dataset: most holonyms in BLESS are rather specific words. 

BLESS was built starting from 200 basic level concepts (e.g. \emph{goldfish}) used as the $x$ words, to which $y$ words in different relations were associated (e.g. \emph{eye}, for meronymy; \emph{animal}, for hypernymy). $x$ words represent hyponyms in the hyponym-hypernym pairs, and should therefore not be too general. Indeed, SLQS assigns high scores to hyponym-hypernym pairs. At the same time, in the meronymy relation in BLESS, $x$ is the holonym and $y$ is the meronym. For consistency with EVALution, we switched those pairs in BLESS, placing the meronym in the $x$ slot and the holonym in the $y$ slot. As a consequence, after the switching, holonyms in BLESS are usually rather specific words (e.g., there are no holonyms like \emph{animal} and \emph{vehicle}, as these words were originally in the $y$ slot). In most cases, they are not more general than their meronyms (\emph{(eye, goldfish)}), yielding low SLQS scores which are easy to separate from hypernyms. We note that this is a weakness of the BLESS dataset, rather than a strength of the measure. For instance, on EVALution, SLQS performs worse (ranked only as high as 13th), as this dataset has no such restriction on the basic level concepts, and may contain pairs like \emph{(eye, animal)}.

\paragraph{Hypernym vs. Attribute} Symmetric similarity measures computed on syntactic contexts succeed to discriminate between hypernyms and attributes. Since attributes are syntactically different from hypernyms (in attributes, $y$ is an adjective), it is unsurprising that they occur in different syntactic contexts, yielding low similarity scores.

\paragraph{Hypernym vs. Antonym} In all our experiments, antonyms were the hardest to distinguish from hypernyms, yielding the lowest performance. We found that SLQS performed reasonably well in this setting. However, the measure variations, context types and feature weightings were not consistent across datasets. SLQS relies on the assumption that $y$ is a more general word than $x$, which is not true for antonyms, making it the most suitable measure for this setting.

\paragraph{Hypernym vs. Synonym} SLQS performs well also in discriminating between hypernyms and synonyms, in which $y$ is also not more general than $x$. 
We observed that in the \smalltt{joint} context type, the difference in SLQS scores between synonyms and hypernyms was the largest.
This may stem from the restrictiveness of this context type. For instance, among the most salient contexts we would expect to find informative contexts like \emph{drinks milk} for \emph{cat} and less informative ones like \emph{drinks water} for \emph{animal}, whereas the non-restrictive single dependency context \emph{drinks} would probably be present for both. 

Another measure that works well is invCL: interestingly, other inclusion-based measures assign high scores to $(x,y)$ when $y$ includes many of $x$'s contexts, which might be true also for synonyms (e.g. \emph{elevator} and \emph{lift} share many contexts). invCL, on the other hand, reduces with the ratio of $y$'s contexts included in $x$, yielding lower scores for synonyms.

\paragraph{Hypernym vs. Coordination} We found no consistency among BLESS and Weeds. On Weeds, inclusion-based measures (ClarkeDE, invCL and Weeds) showed the best results. The best performing measures on BLESS, however, were variants of SLQS, that showed to perform well in cases where the negative relation is symmetric (antonym, synonym and coordination). The difference could be explained by the nature of the datasets: the BLESS test set contains 1,185 hypernymy pairs, with only 129 distinct $y$s, many of which are general words like \emph{animal} and \emph{object}. The Weeds test set, on the other hand, was intentionally constructed to contain an overall unique $y$ in each pair, and therefore contains much more specific $y$s (e.g. \emph{(quirk, strangeness)}). For this reason, generality-based measures perform well on BLESS, and struggle with Weeds, which is handled better using inclusion-based measures.

\subsection{Comparison to State-of-the-art Supervised Methods}
\label{sec:embeddings}

For comparison with the state-of-the-art, we evaluated several supervised hypernymy detection methods, based on the word embeddings of $x$ and $y$: concatenation $\vec{v}_x \oplus \vec{v}_y$ \cite{baroni2012entailment}, difference $\vec{v}_y - \vec{v}_x$ \cite{weeds2014learning}, and ASYM \cite{roller2014inclusive}. We downloaded several pre-trained embeddings \cite{mikolov2013distributed,pennington2014glove,levy2014dependency}, and trained a logistic regression classifier to predict hypernymy. We used the 90\% portion (originally the test set) as the train set, and the other 10\% (originally the validation set) as a test set, reporting the best results among different vectors, method and regularization factor.\footnote{In our preliminary experiments we also trained other classifiers used in the distributional hypernymy detection literature (SVM and SVM+RBF kernel), that performed similarly. We report the results for logistic regression, since we use the prediction probabilities to measure average precision.}

\begin{table*}
	\center
	\small
	\begin{tabular}{ c | c | c | c | c |}
		\hhline{~----}
		& \textbf{method} & \specialcell{\textbf{AP@100} \textbf{original}} & \specialcell{\textbf{AP@100} \textbf{switched}} & $\mathbf\Delta$ \\ \hline
		\multicolumn{1}{|c|}{\textbf{supervised}} & concat, word2vec, $L_1$	& 0.995	& 0.575 & -0.42 \\ \hline
		\multicolumn{1}{|c|}{\textbf{unsupervised}} & cosWeeds, win2d, ppmi & 0.818 & 0.882 & +0.064 \\ \hline
	\end{tabular}
	\vspace*{-5pt}
	\caption{Average Precision (AP) at $k=100$ of the best supervised and unsupervised methods for hypernym vs. random-n, on the original BLESS validation set and the validation set with the artificially added switched hypernym pairs.}
	\label{tab:switched_pairs}
	\vspace*{-10pt}
\end{table*}

Table~\ref{tab:validation_results} displays the performance of the best classifier on each dataset, in a hypernym vs. a single relation setting. We also re-evaluated the unsupervised measures, this time reporting the results on the validation set (10\%) for comparison. 

The overall performance of the embedding-based classifiers is almost perfect, and in particular the best performance is achieved using the concatenation method \cite{baroni2012entailment} with either GloVe \cite{pennington2014glove} or the dependency-based embeddings \cite{levy2014dependency}. As expected, the unsupervised measures perform worse than the embedding-based classifiers, though generally not bad on their own.

These results may suggest that unsupervised methods should be preferred only when no training data is available, leaving all the other cases to supervised methods. This is, however, not completely true. As others previously noticed, supervised methods do not actually learn the relation between $x$ and $y$, but rather separate properties of either $x$ or $y$. \newcite{levy2015supervised} named this the ``lexical memorization'' effect, i.e. memorizing that certain $y$s tend to appear in many positive pairs (\emph{prototypical hypernyms}). 

On that account, the Weeds dataset has been designed to avoid such memorization, with every word occurring once in each slot of the relation. 
While the performance of the supervised methods on this dataset is substantially lower than their performance on other datasets, it is yet well above the random baseline which we might expect from a method that can only memorize words it has seen during training.\footnote{The dataset is balanced between its two classes.} This is an indication that supervised methods can abstract away from the words. 

Indeed, when we repeated the experiment with a lexical split of each dataset, i.e., such that the train and test set consist of distinct vocabularies, we found that the supervised methods' performance did not decrease dramatically, in contrast to the findings of \newcite{levy2015supervised}. 
The large performance gaps reported by \newcite{levy2015supervised} might be attributed to the size of their training sets. Their lexical split discarded around half of the pairs in the dataset and split the rest of the pairs equally to train and test, resulting in a relatively small train set. We performed the split such that only around 30\% of the pairs in each dataset were discarded, and split the train and test sets with a ratio of roughly 90/10\%, obtaining large enough train sets.

Our experiment suggests that rather than memorizing the verbatim \emph{prototypical hypernyms}, the supervised models might learn that certain regions in the vector space pertain to prototypical hypernyms. For example, \emph{device} (from the BLESS train set) and \emph{appliance} (from the BLESS test set) are two similar words, which are both prototypical hypernyms. Another interesting observation was recently made by \newcite{roller2016relations}: they showed that when dependency-based embeddings are used, supervised distributional methods trace $x$ and $y$'s separate occurrences in different slots of Hearst patterns \cite{hearst1992automatic}.

Whether supervised methods only memorize or also learn, it is more consensual that they lack the ability to capture the relation between $x$ and $y$, and that they rather indicate how likely $y$ ($x$) is to be a hypernym (hyponym) \cite{levy2015supervised,santus2016nine,shwartz2016improving,roller2016relations}. While this information is valuable, it cannot be solely relied upon for classification. 

To better understand the extent of this limitation, we conducted an experiment in a similar manner to the switched hypernym pairs in \newcite{santus2016nine}. We used BLESS, which is the only dataset with random pairs. For each hypernym pair $(x_1,y_1)$, we sampled a word $y_2$ that participates in another hypernym pair $(x_2,y_2)$, such that $(x_1,y_2)$ is not in the dataset, and added $(x_1,y_2)$ as a random pair. We added 139 new pairs to the validation set, such as \emph{(rifle, animal)} and \emph{(salmon, weapon)}. We then used the best supervised and unsupervised methods for hypernym vs. random-n on BLESS to re-classify the revised validation set. Table~\ref{tab:switched_pairs} displays the experiment results. 

The switched hypernym experiment paints a much less optimistic picture of the embeddings' actual performance, with a drop of 42 points in average precision. 121 out of the 139 switched hypernym pairs were falsely classified as hypernyms. Examining the $y$ words of these pairs reveals general words that appear in many hypernym pairs (e.g. \emph{animal, object, vehicle}).  The unsupervised measure was not similarly affected by the switched pairs, and the performance even slightly increased. This result is not surprising, since most unsupervised measures aim to capture aspects of the relation between $x$ and $y$, while not relying on information about one of the words in the pair.\footnote{\newcite{turney2015experiments} have also shown that unsupervised methods are more robust than supervised ones in a transfer-learning experiment, when the ``training data'' was used to tune their parameters.}

\section{Discussion}
\label{sec:discussion}

The results in Section~\ref{sec:experiments} suggest that a supervised method using the unsupervised measures as features could possibly be the best of both worlds. We would expect it to be more robust than embedding-based methods on the one hand, while being more informative than any single unsupervised measure on the other hand. 

Such a method was developed by \newcite{santus2016nine}, however using mostly features that describe a single word, e.g. frequency and entropy. It was shown to be competitive with the state-of-the-art supervised methods. With that said, it was also shown to be sensitive to the distribution of training examples in a specific dataset, like the embedding-based methods. 

We conducted a similar experiment, with a much larger number of unsupervised features, namely the various measure scores, and encountered the same issue. While the performance was good, it dropped dramatically when the model was tested on a different test set.

We conjecture that the problem stems from the currently available datasets, which are all somewhat artificial and biased. Supervised methods which are strongly based on the relation between the words, e.g. those that rely on path-based information \cite{shwartz2016improving}, manage to overcome the bias. Distributional methods, on the other hand, are based on a weaker notion of the relation between words, hence are more prone to overfit the distribution of training instances in a specific dataset. In the future, we hope that new datasets will be available for the task, which would be drawn from corpora and will reflect more realistic distributions of words and semantic relations.

\section{Conclusion}

We performed an extensive evaluation of unsupervised methods for discriminating hypernyms from other semantic relations. We found that there is no single combination of measure and parameters which is always preferred; however, we suggested a principled linguistic-based analysis of the most suitable measure for each task that yields consistent performance across different datasets. 

We investigated several new variants of existing methods, and found that some variants of SLQS turned out to be superior on certain tasks. In addition, we have tested for the first time the \smalltt{joint} context type \cite{chersoni2016joint}, which was found to be very discriminative, and might hopefully benefit other semantic tasks.

For comparison, we evaluated the state-of-the-art supervised methods on the datasets, and they have shown to outperform the unsupervised ones, while also being efficient and easier to use. However, a deeper analysis of their performance demonstrated that, as previously suggested, these methods do not capture the relation between $x$ and $y$, but rather indicate the ``prior probability'' of either word to be a hyponym or a hypernym. As a consequence, supervised methods are sensitive to the distribution of examples in a particular dataset, making them less reliable for real-world applications. Being motivated by linguistic hypotheses, and independent from training data, unsupervised measures were shown to be more robust. In this sense, unsupervised methods can still play a relevant role, especially if combined with supervised methods, in the decision whether the relation holds or not.

\section*{Acknowledgments}

The authors would like to thank Ido Dagan, Alessandro Lenci, and Yuji Matsumoto for their help and advice.
Vered Shwartz is partially supported by an Intel ICRI-CI grant, the Israel Science Foundation grant 880/12, and the German Research Foundation through the German-Israeli Project Cooperation (DIP, grant DA 1600/1-1).
Enrico Santus is partially supported by HK PhD Fellowship Scheme under PF12-13656.

\bibliography{hypernymy}

\begin{thebibliography}{}

\bibitem[\protect\citename{Baroni and Lenci}2010]{baroni2010distributional}
Marco Baroni and Alessandro Lenci.
\newblock 2010.
\newblock Distributional memory: A general framework for corpus-based
  semantics.
\newblock {\em Computational Linguistics, Volume 36, Issue 4 - December 2010}.

\bibitem[\protect\citename{Baroni and Lenci}2011]{baroni2011we}
Marco Baroni and Alessandro Lenci.
\newblock 2011.
\newblock Proceedings of the gems 2011 workshop on geometrical models of
  natural language semantics.
\newblock pages 1--10. Association for Computational Linguistics.

\bibitem[\protect\citename{Baroni \bgroup et al.\egroup }2009]{baroni2009wacky}
Marco Baroni, Silvia Bernardini, Adriano Ferraresi, and Eros Zanchetta.
\newblock 2009.
\newblock The wacky wide web: a collection of very large linguistically
  processed web-crawled corpora.
\newblock {\em LREC}.

\bibitem[\protect\citename{Baroni \bgroup et al.\egroup
  }2012]{baroni2012entailment}
Marco Baroni, Raffaella Bernardi, Ngoc-Quynh Do, and Chung-chieh Shan.
\newblock 2012.
\newblock Entailment above the word level in distributional semantics.
\newblock In {\em Proceedings of the 13th Conference of the European Chapter of
  the Association for Computational Linguistics}, pages 23--32. Association for
  Computational Linguistics.

\bibitem[\protect\citename{Benotto}2015]{benotto2015distributional}
Giulia Benotto.
\newblock 2015.
\newblock Distributional models for semantic relations: A sudy on hyponymy and
  antonymy.
\newblock {\em PhD Thesis, University of Pisa}.

\bibitem[\protect\citename{Bullinaria and Levy}2007]{bullinaria2007extracting}
John~A. Bullinaria and Joseph~P. Levy.
\newblock 2007.
\newblock Extracting semantic representations from word co-occurrence
  statistics: A computational study.
\newblock {\em Behavior Research Methods}.

\bibitem[\protect\citename{Chersoni \bgroup et al.\egroup
  }2016]{chersoni2016joint}
Emmanuele Chersoni, Enrico Santus, Alessandro Lenci, Philippe Blache, and
  Chu-Ren Huang.
\newblock 2016.
\newblock Representing verbs with rich contexts: an evaluation on verb
  similarity.
\newblock pages 1967--1972, November.

\bibitem[\protect\citename{Church and Hanks}1989]{church1990word}
Kenneth~Ward Church and Patrick Hanks.
\newblock 1989.
\newblock Word association norms, mutual information, and lexicography.

\bibitem[\protect\citename{Clarke}2009]{clarke2009context}
Daoud Clarke.
\newblock 2009.
\newblock Proceedings of the workshop on geometrical models of natural language
  semantics.
\newblock pages 112--119. Association for Computational Linguistics.

\bibitem[\protect\citename{Dagan \bgroup et al.\egroup
  }2013]{dagan2013recognizing}
Ido Dagan, Dan Roth, and Mark Sammons.
\newblock 2013.
\newblock Recognizing textual entailment.

\bibitem[\protect\citename{Evert}2005]{evert2005statistics}
Stefan Evert.
\newblock 2005.
\newblock The statistics of word cooccurrences: word pairs and collocations.
\newblock {\em Dissertation}.

\bibitem[\protect\citename{Evert}2008]{evert2008corpora}
Stefan Evert.
\newblock 2008.
\newblock Corpora and collocations.
\newblock {\em Corpus linguistics. An international handbook}.

\bibitem[\protect\citename{Ferraresi}2007]{ferraresi2007building}
Adriano Ferraresi.
\newblock 2007.
\newblock Building a very large corpus of english obtained by web crawling:
  ukwac.
\newblock {\em Master’s thesis, University of Bologna, Italy}.

\bibitem[\protect\citename{Harris}1954]{harris1954distributional}
Zellig~S. Harris.
\newblock 1954.
\newblock Distributional structure.
\newblock {\em Word}.

\bibitem[\protect\citename{Hearst}1992]{hearst1992automatic}
Marti~A. Hearst.
\newblock 1992.
\newblock Automatic acquisition of hyponyms from large text corpora.
\newblock In {\em COLING 1992 Volume 2: The 15th International Conference on
  Computational Linguistics}.

\bibitem[\protect\citename{Kiela \bgroup et al.\egroup
  }2015]{kiela2015exploiting}
Douwe Kiela, Laura Rimell, Ivan Vuli{\'{c}}, and Stephen Clark.
\newblock 2015.
\newblock Exploiting image generality for lexical entailment detection.
\newblock In {\em Proceedings of the 53rd Annual Meeting of the Association for
  Computational Linguistics and the 7th International Joint Conference on
  Natural Language Processing (Volume 2: Short Papers)}, pages 119--124.
  Association for Computational Linguistics.

\bibitem[\protect\citename{Kotlerman \bgroup et al.\egroup
  }2010]{kotlerman2010directional}
Lili Kotlerman, Ido Dagan, Idan Szpektor, and Maayan Zhitomirsky-Geffet.
\newblock 2010.
\newblock Directional distributional similarity for lexical inference.
\newblock {\em NLE}.

\bibitem[\protect\citename{Lenci and Benotto}2012]{lenci2012identifying}
Alessandro Lenci and Giulia Benotto.
\newblock 2012.
\newblock Identifying hypernyms in distributional semantic spaces.
\newblock In {\em *SEM 2012: The First Joint Conference on Lexical and
  Computational Semantics -- Volume 1: Proceedings of the main conference and
  the shared task, and Volume 2: Proceedings of the Sixth International
  Workshop on Semantic Evaluation (SemEval 2012)}, pages 75--79. Association
  for Computational Linguistics.

\bibitem[\protect\citename{Levy and Goldberg}2014]{levy2014dependency}
Omer Levy and Yoav Goldberg.
\newblock 2014.
\newblock Dependency-based word embeddings.
\newblock In {\em Proceedings of the 52nd Annual Meeting of the Association for
  Computational Linguistics (Volume 2: Short Papers)}, pages 302--308.
  Association for Computational Linguistics.

\bibitem[\protect\citename{Levy \bgroup et al.\egroup
  }2015a]{levy2015improving}
Omer Levy, Yoav Goldberg, and Ido Dagan.
\newblock 2015a.
\newblock Improving distributional similarity with lessons learned from word
  embeddings.
\newblock {\em Transactions of the Association of Computational Linguistics --
  Volume 3, Issue 1}, pages 211--225.

\bibitem[\protect\citename{Levy \bgroup et al.\egroup
  }2015b]{levy2015supervised}
Omer Levy, Steffen Remus, Chris Biemann, and Ido Dagan.
\newblock 2015b.
\newblock Do supervised distributional methods really learn lexical inference
  relations?
\newblock pages 970--976.

\bibitem[\protect\citename{Lin}1998]{lin1998information}
Dekang Lin.
\newblock 1998.
\newblock An information-theoretic definition of similarity.
\newblock In {\em ICML}.

\bibitem[\protect\citename{Mikolov \bgroup et al.\egroup
  }2013]{mikolov2013distributed}
Tomas Mikolov, Ilya Sutskever, Kai Chen, Gregory~S. Corrado, and Jeffrey Dean.
\newblock 2013.
\newblock Distributed representations of words and phrases and their
  compositionality.
\newblock In {\em NIPS}.

\bibitem[\protect\citename{Navigli \bgroup et al.\egroup
  }2011]{navigli2011graph}
Roberto Navigli, Paola Velardi, and Stefano Faralli.
\newblock 2011.
\newblock A graph-based algorithm for inducing lexical taxonomies from scratch.
\newblock In {\em IJCAI}.

\bibitem[\protect\citename{Pad{\'o} and Lapata}2007]{pado2007dependency}
Sebastian Pad{\'o} and Mirella Lapata.
\newblock 2007.
\newblock Dependency-based construction of semantic space models.
\newblock {\em Computational Linguistics, Volume 33, Number 2, June 2007}.

\bibitem[\protect\citename{Pennington \bgroup et al.\egroup
  }2014]{pennington2014glove}
Jeffrey Pennington, Richard Socher, and Christopher Manning.
\newblock 2014.
\newblock Glove: Global vectors for word representation.
\newblock In {\em Proceedings of the 2014 Conference on Empirical Methods in
  Natural Language Processing (EMNLP)}, pages 1532--1543. Association for
  Computational Linguistics.

\bibitem[\protect\citename{Rimell}2014]{rimell2014distributional}
Laura Rimell.
\newblock 2014.
\newblock Distributional lexical entailment by topic coherence.
\newblock In {\em Proceedings of the 14th Conference of the European Chapter of
  the Association for Computational Linguistics}, pages 511--519. Association
  for Computational Linguistics.

\bibitem[\protect\citename{Roller and Erk}2016]{roller2016relations}
Stephen Roller and Katrin Erk.
\newblock 2016.
\newblock Relations such as hypernymy: Identifying and exploiting hearst
  patterns in distributional vectors for lexical entailment.
\newblock pages 2163--2172, November.

\bibitem[\protect\citename{Roller \bgroup et al.\egroup
  }2014]{roller2014inclusive}
Stephen Roller, Katrin Erk, and Gemma Boleda.
\newblock 2014.
\newblock Inclusive yet selective: Supervised distributional hypernymy
  detection.
\newblock In {\em Proceedings of COLING 2014, the 25th International Conference
  on Computational Linguistics: Technical Papers}, pages 1025--1036. Dublin
  City University and Association for Computational Linguistics.

\bibitem[\protect\citename{Salton and McGill}1986]{salton1986introduction}
Gerard Salton and Michael~J. McGill.
\newblock 1986.
\newblock Introduction to modern information retrieval.

\bibitem[\protect\citename{Santus \bgroup et al.\egroup
  }2014]{santus2014chasing}
Enrico Santus, Alessandro Lenci, Qin Lu, and Sabine Schulte~im Walde.
\newblock 2014.
\newblock Chasing hypernyms in vector spaces with entropy.
\newblock In {\em Proceedings of the 14th Conference of the European Chapter of
  the Association for Computational Linguistics, volume 2: Short Papers}, pages
  38--42. Association for Computational Linguistics.

\bibitem[\protect\citename{Santus \bgroup et al.\egroup
  }2015]{santus2015evalution}
Enrico Santus, Frances Yung, Alessandro Lenci, and Chu-Ren Huang.
\newblock 2015.
\newblock Proceedings of the 4th workshop on linked data in linguistics:
  Resources and applications.
\newblock pages 64--69. Association for Computational Linguistics.

\bibitem[\protect\citename{Santus \bgroup et al.\egroup
  }2016a]{santus2016unsupervised}
Enrico Santus, Tin-Shing Chiu, Qin Lu, Alessandro Lenci, and Chu-Ren Huang.
\newblock 2016a.
\newblock Unsupervised measure of word similarity: How to outperform
  co-occurrence and vector cosine in vsms.
\newblock {\em AAAI}.

\bibitem[\protect\citename{Santus \bgroup et al.\egroup }2016b]{santus2016nine}
Enrico Santus, Alessandro Lenci, Tin-Shing Chiu, Qin Lu, and Chu-Ren Huang.
\newblock 2016b.
\newblock Nine features in a random forest to learn taxonomical semantic
  relations.
\newblock {\em LREC}.

\bibitem[\protect\citename{Shwartz \bgroup et al.\egroup
  }2016]{shwartz2016improving}
Vered Shwartz, Yoav Goldberg, and Ido Dagan.
\newblock 2016.
\newblock Improving hypernymy detection with an integrated path-based and
  distributional method.
\newblock In {\em Proceedings of the 54th Annual Meeting of the Association for
  Computational Linguistics (Volume 1: Long Papers)}, pages 2389--2398.
  Association for Computational Linguistics.

\bibitem[\protect\citename{Snow \bgroup et al.\egroup }2004]{snow2004learning}
Rion Snow, Daniel Jurafsky, and Andrew~Y. Ng.
\newblock 2004.
\newblock Learning syntactic patterns for automatic hypernym discovery.
\newblock In {\em NIPS}.

\bibitem[\protect\citename{Snow \bgroup et al.\egroup }2006]{snow2006semantic}
Rion Snow, Daniel Jurafsky, and Andrew~Y. Ng.
\newblock 2006.
\newblock Semantic taxonomy induction from heterogenous evidence.
\newblock In {\em Proceedings of the 21st International Conference on
  Computational Linguistics and 44th Annual Meeting of the Association for
  Computational Linguistics}, pages 801--808. Association for Computational
  Linguistics.

\bibitem[\protect\citename{Turney and Mohammad}2015]{turney2015experiments}
Peter~D. Turney and Saif~M. Mohammad.
\newblock 2015.
\newblock Experiments with three approaches to recognizing lexical entailment.
\newblock {\em Natural Language Engineering}, 21(03):437--476.

\bibitem[\protect\citename{Weeds and Weir}2003]{weeds2003general}
Julie Weeds and David Weir.
\newblock 2003.
\newblock Proceedings of the 2003 conference on empirical methods in natural
  language processing.

\bibitem[\protect\citename{Weeds \bgroup et al.\egroup
  }2014]{weeds2014learning}
Julie Weeds, Daoud Clarke, Jeremy Reffin, David Weir, and Bill Keller.
\newblock 2014.
\newblock Learning to distinguish hypernyms and co-hyponyms.
\newblock In {\em Proceedings of COLING 2014, the 25th International Conference
  on Computational Linguistics: Technical Papers}, pages 2249--2259. Dublin
  City University and Association for Computational Linguistics.

\end{thebibliography}
\bibliographystyle{eacl2017}

\end{document}